# A Low Cost Vision Based Hybrid Fiducial Mark Tracking Technique for Mobile Industrial Robots

Mohammed Y Aalsalem[1], Wazir Zada Khan[2] and Quratul Ain Arshad[3]

[1] School of Computer Science, University of Jazan
Jazan, PoBox # 114, Kingdom of Saudi Arabia
*aalsalem.m@jazanu.edu.sa*

[2] School of Computer Science, University of Jazan
Jazan, PoBox # 114, Kingdom of Saudi Arabia
*wazirzadakhan@jazanu.edu.sa*

[3] School of Computer Science, University of Jazan
Jazan, PoBox # 114, Kingdom of Saudi Arabia
*brightsuccess_12@yahoo.com*

**Abstract**
The field of robotic vision is developing rapidly. Robots can react intelligently and provide assistance to user activities through sentient computing. Since industrial applications pose complex requirements that cannot be handled by humans, an efficient low cost and robust technique is required for the tracking of mobile industrial robots. The existing sensor based techniques for mobile robot tracking are expensive and complex to deploy, configure and maintain. Also some of them demand dedicated and often expensive hardware. This paper presents a low cost vision based technique called "Hybrid Fiducial Mark Tracking" (HFMT) technique for tracking mobile industrial robot. HFMT technique requires off-the-shelf hardware (CCD cameras) and printable 2-D circular marks used as fiducials for tracking a mobile industrial robot on a pre-defined path. This proposed technique allows the robot to track on a predefined path by using fiducials for the detection of Right and Left turns on the path and White Strip for tracking the path. The HFMT technique is implemented and tested on an indoor mobile robot at our laboratory. Experimental results from robot navigating in real environments have confirmed that our approach is simple and robust and can be adopted in any hostile industrial environment where humans are unable to work.

**Keywords:** *Context-Aware Computing, Mobile Robots Path Tracking, Fiducial Detection, Computer Vision, Autonomous Robot Navigation.*

## 1. Introduction

Today, robotics is a rapidly growing field, as we continue to research, design, and build new robots that dole out various practical purposes, whether domestically, commercially, or militarily [1]. Robots have revolutionized the industrial workplace. Robots are now gearing up to play a major role in less structured and more flexible manufacturing environments. Thousands of manufacturers are dependant upon the productivity, high performance and savings provided by modern-day industrial automation. Robots range from humanoids such as ASIMO and TOPIO to Nano robots, Swarm robots, Industrial robots, military robots, mobile and servicing robots. Robots offer specific benefits to workers, industries and countries. Robots can be used for a wide range of industrial applications as industrial robots are the effective application of the modern robotics. If introduced correctly, industrial robots can improve the quality of life by freeing workers from dirty, boring, dangerous and heavy labor. It is true that robots can cause unemployment by replacing human workers, but robots also create jobs for robot technicians, salesmen, engineers, programmers, and supervisors.

An industrial robot is defined by ISO [4] as "an automatically controlled, reprogrammable, multipurpose manipulator programmable in three or more axes". The field of robotics may be more practically defined as the study, design and use of robot systems for manufacturing. George Charles Devol is often called the father of robotics. He invented the first industrial robot, the Unimate, in 1954. A few years later, Devol and Joseph F Engelberger formed the first robot company, Unimation. In 1960, Unimation was purchased by Condec Corporation. General Motors installed the Unimate for die casting handling and spot welding in 1961 [3]. Modern industrial robots are true marvels of engineering. A robot the size of a person can easily carry a load over one hundred pounds and move it very quickly with a repeatability of +/- 0.006 inches. Furthermore these robots can do that 24 hours a day for years on end with no failures whatsoever. Though they are reprogrammable, in many applications they are programmed once and then repeat that exact same task for years [2]. In early times, robots were initially applied to jobs that were hot, heavy and hazardous like die casting, forging, spot welding, and spray painting. Nowadays modern industrial robots can be easily programmed to perform new applications.

Industrial robotics plays a key role in automation. It has improved manufacturing processes, performing functions precisely and quickly because robots out perform humans in

jobs that require precision, speed, endurance and reliability. The advantages of robotics are as follows:
1. Many industrial robots are in the form of a robotic arm. The first industrial robot called Unimate, had the appearance of a robotic arm. Due to its mechanical nature and computerized control, a robotic arm can carry out a repetitive task with great precision and accuracy, thus providing improved, consistent product quality. This would apply to quite a variety of production line tasks, like welding, assembling a product, spray painting, or cutting and finishing.
2. The two main features of industrial robotics i.e., the mechanical nature of the equipment and the computerized control make industrial robotics technology more efficient and speedy, leading to higher production rates than with human labor. Another aspect of efficiency is that robots can be mounted from the ceiling and have no problem with working upside down. This can lead to a savings in floor space.
3. There are a number of tasks that are too dangerous, too exposed to toxins, or just plain too dirty for humans to conveniently do them. These are ideal robotics tasks. This includes tasks as simple as spray painting, because there is no need to worry about the robot inhaling the paint fumes. It also includes such daunting tasks as defusing bombs and such dirty tasks as cleaning sewers. Thus robots can be used effectively in such environments where handling of radioactive materials is involved, such as hospitals or nuclear establishments, where direct exposure to human beings can be dangerous for their health.
4. Another advantage of robotics is due to the fact that human characteristics like boredom from doing a repetitive task don't interfere with the functioning of a robot. In many production establishments work required to be executed is awfully boring, being cyclic and repetitive, due to which it is difficult for the operators to remain fully dedicated to their tasks and generate interest in their work. When tasks are monotonous, workers tend to be careless, thereby increasing the probability of accidents and malfunctions of machines. Utilization of robots has eliminated problems associated with boredom in production. Since a robot doesn't need to rest or eat, and never gets sick, a robotic arm can work twenty-four hours, during day or night, on holidays, without any break so as to ensure increased production, with only limited occasional downtime for scheduled maintenance.

Machine vision is the ability of a computer to "see." A machine-vision system employs one or more video cameras, analog-to-digital conversion (ADC), and digital signal processing (DSP). The resulting data goes to a computer or robot controller [7]. Machine Vision gives robots intelligent eyes. Using these eyes, robots can recognize the position of objects in space and adjust their working steps accordingly.

Industrial robot vision systems can be utilized in a variety of ways [5]. From reducing downtime by avoiding collisions to checking parts for defects, industrial robotic systems provide efficiency, safety, and savings and productivity. Industrial robotic vision systems integrate complex algorithms, calibration, temperature software, and cameras together to identify parts, orient parts for robotic handling, navigate work envelopes, and find scattered parts. Robot vision is used for part identification and navigation. These systems can also be used to improve welding techniques. Weld seam tracking performed by welding robots with vision systems integrated results in less rework, less waste, and better welds. A robot with a vision system can also perform the tasks faster and with better repeatability than a human.

The Robot Vision is comprised of technologies like Fast and precise position determination of points, features or bodies in the plane or space using robust algorithms, Coordinate transformations between the coordinate systems of the camera, the robot, the object and the cell and the calibration of the entire system, Time-efficient communication, resistant to transmission errors, between the image processor and the robot via various interfaces and protocols, Integration of additional procedures for quality measurement and inspection including the documentation of all relevant results, and Clear structuring of the software and concise user-interface in order to make complex technology available in a simple manner.

The main goal of Robot Vision is to design systems for automating automation - everything from simple pick & place applications to complex 3D tasks and implement these cost-effectively [6]. The demands on Robot Vision Systems start at the two-dimensional recognition of the translation and rotation of an object on a conveyor belt and extend beyond container unloading, to the recognition of all 6 degrees of freedom of the position of a moving object in space. The 2D robotic vision systems consist of standard industrial cameras used to take images that are processed by the robot to make decisions on how parts should be handled. Some of the applications of 2D vision systems are Positioning Pick & Place systems, loading and unloading processing machines, sorting from the conveyor belt, loading and unloading latticed boxes, filling barrels, controlling sanding and milling robots, Guiding harvesting machines etc. A 2D robotic vision is shown in figure.

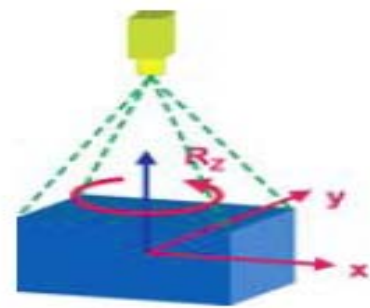

Figure 1: 2D Robot Vision [6]

Real-time vision tasks for small autonomous robots include object tracking, obstacle detection and avoidance, and path planning. The main requirement of augmented reality is accurate indoor and outdoor position or location tracking and orientation tracking, thus localization is a major concerning problem for an autonomous vehicle such as robot. A significant research effort has been made and different techniques have been developed so far in order to deal with this localization problem. These existing techniques are broadly categorized into two types, Relative (local) Localization and Absolute (Global) Localization [9]. In Relative Localization the position and orientation is evaluated through integration of information provided by diverse sensors. The integration is started from the initial position and is continuously updated in time. Local Localization is also known as *Dead-Reckoning* (DR). Absolute Localization consists of those techniques which allow the vehicle to find its way directly in the domain of evolution of mobile system. These techniques usually rely on navigation beacons, active or passive landmarks, map matching or satellite based signals like GPS. Although DR methods are simple but we cannot rely on these for long distances. As DR methods or Local Localization techniques use sensors, they offer many disadvantages. First, the sensor models suffer from inaccuracies and can become very complicated. Second, the senor readings are corrupted by noise. Third, the motion of the robot involves external sources of error that are not observable by the sensors used for example, slippage in the direction of motion or in the perpendicular direction. Thus due to the above reasons, there is error in the calculation of the robot's position and orientation which generally grows unbounded with time [9]. Another obvious disadvantage of such sensor-based methods is that the robot needs to sense accurate data in the unknown environment [8]. Many Absolute or Global Localization techniques have been developed but they also suffer from some shortcomings.

In a warehouse or plant, instead of conveyor belts, a mobile robot would be more flexible and could be easier adapted to changes in the environment. Such robots follow a marked path on the floor thus position or orientation tracking on a predefined path is a significant task for such industrial robots. The goal of our research is to develop a simple and robust technique to follow a planned marked path. Our proposed technique called *Hybrid Fiducial Mark Tracking (HFMT)* technique is a hybrid technique which is a combination of two existing path tracking methods, Fiducial Detection and Mobile Robot Path Tracking. We have combined the benefits of both techniques developing a more efficient, robust path tracking method for tracking mobile industrial robot on a predefined path with a need of less computational power and resources.

The rest of the paper is organized as follows. Section 2 describes the existing related work Section 3 presents our proposed technique which is low cost and robust technique for tracking the Mobile robot on a predefined path. Our approach is solely 2D-vision based. Finally Section 4 concludes the paper and also describes our future work.

## 2. Related Work

One pragmatic approach to solving the monocular-vision tracking problem is to place a special target marker, or Fiducial, on the object. These distinctively shaped and colored markers make it easier not only to find the target with simple computer vision algorithms, but also to obtain its position, or even its full pose (depending on the marker used) [11].

The Fiducials are an easy way to identify a particular marker within the current image view [10]. Fiducials are commonly known as "markers that are easy to identify" and are typically added to an environment where localization and navigation are needed for robots using machine vision. Having the fiducial in sight helps the robot to know where it is with respect to the fiducial and can use the fiducial as a target for navigation based on scale, rotation and translation. Fiducials are also typically 2D planar objects that have very distinct corners or shapes. Identifying these traits in the environment helps to reduce the "clutter" of other objects and ensures better identification reliability. Many researchers have proposed such systems and techniques which make use of fiducials for tracking robots in real time environments.

Alan Mutka et al. [12] have explored the possibility to realize an inexpensive and simple navigation system, based on visual feedback, which is able to guide an autonomous mobile robot on a predefined path. They have used low cost embedded platform hardware.Their intention is to add the proposed navigation system to the existing commercial manually driven cleaning vehicles, to turn them into automatic ones. Their proposed system is a visual feedback indoor control system based on the passive visual markers which are placed on the ceiling. Visual markers contain information about a direction angle marker and ID number. Off the shelf webcam, mounted on the top of a mobile robot SHREC (System for Human Replacement in Economical Cleaning), captures RGB images of the ceiling. Once the robot detects the marker, it identifies marker's features and determines a new direction of its movement.

Diego Lopez et al. [13] have proposed a low cost vision based location system called TRIP (Target Recognition using Image Processing) for Ubiquitous Computing. It is Vision based sensor system that sues a combination of 2-D circular barcode tags or ring codes and inexpensive CCD cameras to identify and locate tagged objects in the cameras field of view x

## 3. Proposed *HFMT* Technique

Our proposed technique is a combination of two methods. We have integrated the method of fiducial detection proposed by David et al [14] with the mobile path tracking algorithm proposed by Andrew et al [19]. In [14] the first stage of fiducial detection is to detect candidates in the image. A library of example Fiducials is prepared by capturing video sequences of the markers in a variety of environments. Example Fiducial images are shown in Figure 2, including variation due to foreshortening, motion blur, and lighting.

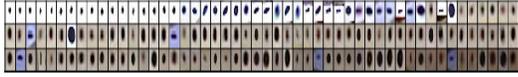

Figure 2: representative samples of positive target images.

In surveying the fiducials, algorithm independently computes the 3D position of each of the four discs on each target. This allows for deviations from planarity in paper targets. In order to determine the position of a particular disc, its 2D positions are gathered in every frame in which it was detected as a list of (x, y, f) tuples, where (x, y) is the marker's position in frame f. Then the 3D point required is that which minimizes the re-projection error. Where the 3D-to-2D perspective projection function is defined as π(x, y, z) = (x/z, y/z). Although this minimization has no closed-form solution, it is readily solved by initialization using the DLT method [15] followed by nonlinear optimization of ∈(X). By repeating this procedure for all fiducials, the accurate estimates of the fiducial positions are obtained. This algorithm provides excellent performance provided that lighting conditions are suitable [1]. However, in difficult lighting conditions, such as outdoors or in low light, the ARToolkit's detection performance drops. In addition, low-resolution images, extreme foreshortening, motion blur, and specular reflections all cause performance to drop. Modifications such as the use of adaptive threshold [16] and homomorphic image processing [17] alleviate some of these problems, but each modification imposes additional computational cost, and introduces an additional failure mode to the algorithm.

In Mobile Path Tracking algorithm the path is indicated by a white line on the playing field [18]. Figure 3 shows two views of the robots camera when tracking a path. If the path is to the right/left, then the robot should turn right/left. Also, if the path is sloping towards the right/left, then the robot should turn to the right/left. A problem occurs if for example the robot is to the right of the path, but the path is sloping towards the left. Therefore, there is a need to combine the offset and the gradient need in order to compute the desired steering behavior.

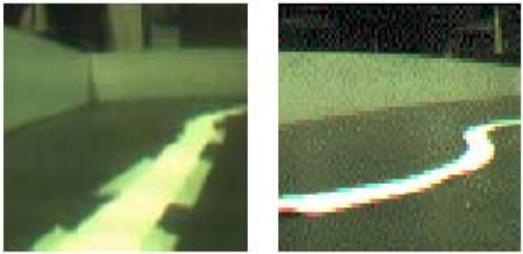

Figure 3: views from the camera of the robot.

Keeping in mind the above mentioned problems with both the Fiducial Detection method and Mobile Robot Path Tracking method, we have proposed a hybrid technique called *"Hybrid Fiducial Mark Tracking" (HFMT)*. In our proposed technique the robot is supposed to follow a marked path using black dots on a white background. The path that the robot will follow is shown in figure 4. We have constructed a path and images (fiducials) are drawn on boards, which are placed on initial and destination positions and on every turn. We have proposed a technique combing the benefits of both methods resulting in a more robust and low cost technique. It is robust against distortion and occlusions which is especially important in populated environments

In our experimental setup, when the robot will find the image by fiducial detection algorithm [14], it compares that image with the pre-defined pattern. If the given markers have three dots arranged in triangle (indicating right turn), or one dot (indicating left turn), and four dots arrange in square, will verify the initial or destination position of the robot. And with the help of image processing and control algorithm [18] the robot will move along the path.

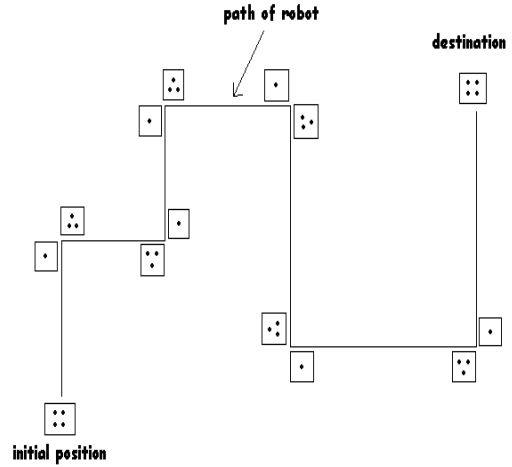

Figure 4: Path for robot to follow

Our proposed HFMT technique solves the problem of lighting conditions in Fiducial Detection by considering predefined paths within the industrial areas where these kinds of conditions can be handled easily. The problem of Mobile Robot Path Tracking method was that the robot suffers from a difficulty of taking right and left turns on a path and consequently moves in wrong directions. Our HFMT technique solves this problem by using specified fiducials.

## 4. Hybrid Fiducial Mark Trcaking Algorithm (HFMT)

Our HFMT algorithm includes the following steps:

1. Movement along the path follows the simple color detection and segmentation routine is used to extract the path from the image.

2. Then the offset $O_p$ and gradient $G_p$ of the path are approximated.

3. On reaching to the board the robot will follow the fiducial detection algorithm given above and detect the fiducial.

---

The proposed *HFMT* algorithm is as under:

   Extract path using simple color detection and segmentation routine

   Calculate offset //to take a start from the initial stage

   Calculate gradient

   Input image I(x, y)

   For i = 1 to 3

   S = f[i] (f-array holds conditions: f[1]=right turn, f[2]=left turn, f[3]=destination)

   If (s == f[1]) //if to turn right

   {

   Turn right; //moving using mobile robot path tracking algorithm

   Calculate offset;

   Calculate gradient;

   }

   If (s == f[2]) // if to turn left

   {

   Turn left; //moving using mobile robot path tracking algorithm

   Calculate offset;

   Calculate gradient;

   }

   If (s == f[3]) ; //if the robot reached to the destination
    Destination;

---

4. Then it will turn right or left according to the detection of the fiducial.

## 5. Evaluation of Proposed Technique

In order to evaluate our proposed technique, we have allowed the robot to track over a predefined path which is shown in figure. For the evaluation we have used a Controlled Robot. The technical details of this robot are shown in Table 1.

In our experiment we have used VEX v.5 Robotics kit. Our robot is four wheeled. The Controlled Robot consists of two VEX Camera Kits, VEX RF Transmitter + Receiver Kit, PIC Microcontroller V0.5, VEX Optical Shaft Encoder. The four wheeled robot and the fiducials used in the lab experiment are shown in Figure 4.

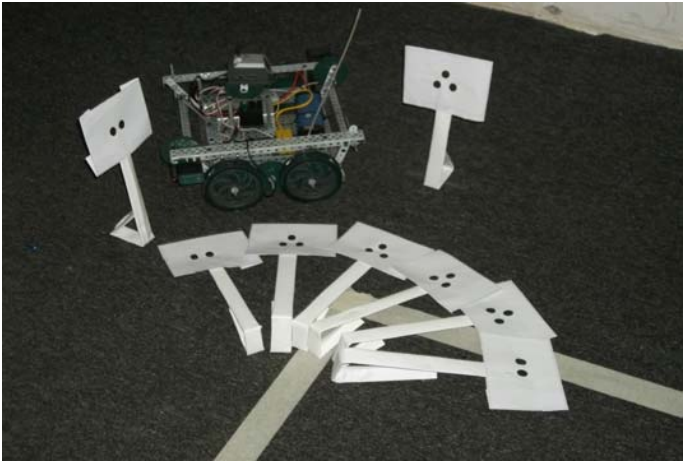

Figure 5: The four wheeled Robot and the fiducials used in the lab experiment

The VEX transmitter is connected to a laptop. The two camera units of two VEX camera kits are mounted on the robot, one on the upper side and one on the lower side of the robot. The camera receivers are connected to the laptop. An external video card is also connected to a laptop and both of the cameras send pictures wirelessly to the external video card. On the laptop we have our main program which takes all the images from the external video card and after processing them, it passes instructions for the robot through the VEX transmitter. This is how the main software controls the robot movement by taking decisions autonomously. Figure 6 shows the testing field and the specified path on which our robot moves using the software and takes decisions on left and right turns on the path. Figure 7 shows the robot moving on the given specified path.

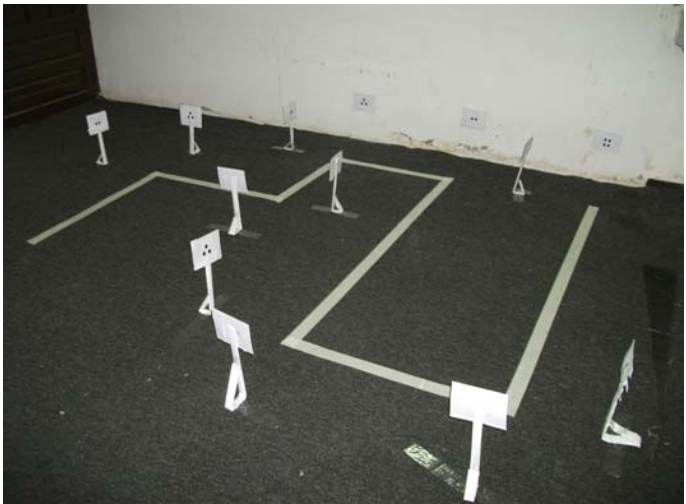

Figure 6: The testing field

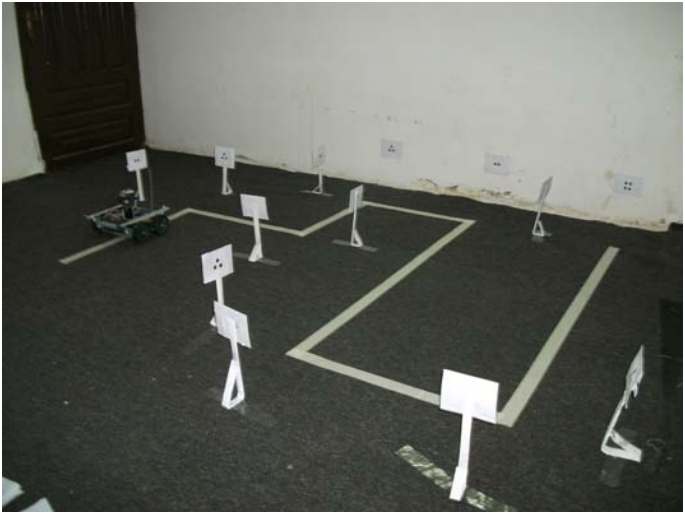

Figure 7: The robot moving on the given specified path.

## 6. Conclusion and Future Work

We have introduced an accurate vision based position and orientation tracking technique named *"Hybrid Fiducial Mark Tracking"* (HFMT) technique for predefined paths used by industrial robots. Our proposed technique combines two algorithms Fiducial Detection and Mobile Robot Path Tracking algorithms, allowing the robot to first detect the path and compare it to the pre-defined pattern that is fed into the memory of the robot. The *HFMT* algorithm is best for the situations where robot has to follow a predefined path. For example: In usual case of industrial applications where robots are used to pick things from any specific location and have to take those things to another location. For the evaluation of our proposed technique we have tested it by allowing the robot to track on a predefined path. We have used the controlled robot for the evaluation of our algorithm. Results have proved that our proposed technique is highly reliable and improves the performance and ease of setup of camera tracking. In future we will handle curves and obstacles that will come on the path of mobile robot. Also our future work will be devoted to the improvement of marker detection capabilities in case of significant change in light conditions with markers positioned on the door and/or on the shelves in a store. Finally, proposed technique shall be implemented using an autonomous robot and will be tested under realistic operating conditions.

A lot of efforts have been made in the field of robotics but still there is a need that researchers and developers must come up with new efficient industrial robots and robotic software and efficient techniques for controlling robot diverse movements in different hostile environments.

**Dr. Muhammad Y Aalsalem** is currently dean of e-learning and assistant professor at School of Computer Science, Jazan University. Kingdom of Saudi Arabia. He received his PhD in Computer Science from Sydney University. His research interests include real time communication, network security, distributed systems, and wireless systems. In particular, he is currently leading in a research group developing flood warning system using real time sensors. He is Program Committee of the International Conference on Computer Applications in Industry and Engineering, CAINE2011. He is regular reviewer for many international journals such as King Saud University Journal (CCIS-KSU Journal).

**Wazir Zada Khan** is currently with School of Computer Science, Jazan University, Kingdom of Saudi Arabia. He received his MS in Computer Science from Comsats Institute of Information Technology, Pakistan. His research interests include network and system security, sensor networks, wireless and ad hoc networks. His subjects of interest include Sensor Networks, Wireless Networks, Network Security and Digital Image Processing, Computer Vision.

**Quratul Ain Arshad** received her BS in Computer Science from Comsats Institute of Information Technology, Pakistan. Her research interests include network security, trust and reputation in sensor networks and ad hoc networks. Her subjects of interest include Network Security, Digital Image Processing and Computer Vision.